\documentclass{article}

\usepackage{arxiv}

\usepackage[utf8]{inputenc} 
\usepackage[T1]{fontenc}    
\usepackage{hyperref}       
\usepackage{url}            
\usepackage{booktabs}       
\usepackage{amsfonts}       
\usepackage{nicefrac}       
\usepackage{microtype}      
\usepackage{lipsum}

\usepackage{graphicx}
\usepackage{epstopdf}
\usepackage{amssymb}
\usepackage{amsmath}
\usepackage{multirow}
\usepackage{color}
\usepackage{tabularx,booktabs}
\usepackage{array}
\newcolumntype{P}[1]{>{\centering\arraybackslash}p{#1}}
\usepackage{booktabs, caption, makecell}

\usepackage{threeparttable}

\usepackage{subcaption}

\title{Fast and Scalable Earth Texture Synthesis\\using Spatially Assembled Generative Adversarial Neural Networks}

\author{
  Sung Eun ~Kim \thanks{saint.kse@gmail.com; https://github.com/saint-kim}\\
  Dept. of Safety and Environment Research\\
  The Seoul Institute\\
  Seoul, 06756 South Korea \\
  \texttt{sekim@si.re.kr} \\
   \And
 Hongkyu ~Yoon \\
  Geomechanics Department\\
  Sandia National Laboratories\\
  Albuquerque, NM, 87123 USA \\
  \texttt{hyoon@sandia.gov} \\
   \And
 Jonghyun ~Lee \\
  Water Resources Research Center \\
  Civil and Environmental Engineering\\
  University of Hawai'i at M\=anoa\\
  Honolulu, HI, 96822 USA \\
  \texttt{jonghyun.harry.lee@hawaii.edu} \\
}

\begin{document}
\maketitle

\begin{abstract}
The earth texture with complex morphological geometry and compositions such as shale and carbonate rocks, is typically characterized with sparse field samples because of an expensive and time-consuming characterization process. 
Accordingly, generating arbitrary large size of the geological texture with similar topological structures at a low computation cost has become one of the key tasks for realistic geomaterial reconstruction.  
Recently, generative adversarial neural networks (GANs) have demonstrated a potential of synthesizing input textural images and creating equiprobable geomaterial images. 
However, the texture synthesis with the GANs framework is often limited by the computational cost and scalability of the output texture size.
In this study, we proposed a spatially assembled GANs (SAGANs) that can generate output images of an arbitrary large size regardless of the size of training images with computational efficiency.
The performance of the SAGANs was evaluated with two and three dimensional (2D and 3D) rock image samples widely used in geostatistical reconstruction of the earth texture.
We demonstrate SAGANs can generate the arbitrary large size of statistical realizations with connectivity and structural properties similar to training images, and also can generate a variety of realizations even on a single training image.
In addition, the computational time was significantly improved compared to standard GANs frameworks. 
\end{abstract}

\keywords{Earth Texture Synthesis \and Deep Learning \and Spatially Assembled Generative Adversarial Networks (SAGANs) \and Microstructures \and Stochastic Modeling }

\section{Introduction}

\label{S:1}
Earth text features such as pore morphology of geomaterials with small, complex pore networks, lithological features of shallow crust, to the internal structure and surface texture of Earth and surface features are crucial for improving our understanding of complex Earth processes. 
In the subsurface where ongoing energy related technologies such as unconventional gas and oil recovery, geologic storage of CO2, and renewable energy recovery are actively optimized and developed, realistic synthesis of textural structures including topology and connectivity is key to improving the prediction accuracy of coupled geological, hydrological, and physical and chemical processes. 
The goal of texture synthesis is often to capture the spatial structural patterns and characteristics from given example images and create many images with similar statistical properties \cite{BergmannJV17}.
However, the earth geomaterials usually have complex textural properties that are often heterogeneous; random characteristics are formally defined as stationary, ergodic, and stochastic processes \cite{BergmannJV17,Georgiadis2013}. 
Sometimes the texture of earth geomaterials with complex topological geometry and compositions such as shale and carbonate rocks usually requires an expensive and time-consuming characterization process and the number of available samples for subsequent subsurface reservoir management design is usually small \cite{tahmasebi2012reconstruction,mariethoz2014bridges}. 
Therefore, accurate capture and realization of the underlying complex stochastic properties of the geological texture with a limited set of samples has long been an important issue in the earth texture synthesis.

A variety of methods have been developed for accurate realization of the geological texture.
For example, instance-based approaches resample pixels or patches of the original texture image, and then copied next to similar image region to generate a seamless bigger texture image \cite{Efros1999,Efros2001,BergmannJV17}.
But, these methods cannot generate the textures with complex spatial patterns due to a limit to account for the spatial variability.
To overcome this limit, spatial covariance or variogram analysis was used to quantify spatial (dis-)similarity and continuity.
These methods have the advantage to quantify the spatial variability in a mathematically tractable way \cite{Goovaerts1998,isaaks1989applied,Li2015}. 
However, the application of these methods with two-point statistics is intrinsically limited to multi-Gaussian systems and/or continuous features, which has been limited to describe high-order statistics and realistic connectivity patterns \cite{Journel1993,Li2015}.

Multiple-point statistics (MPS) was proposed to overcome these problems by adopting image-based approaches where training images (TIs) are used as a basis for sample reconstruction \cite{Guardiano1993}. 
TIs are usually assumed to exhibit the stationarity of the probability distribution of the desired properties and possess higher order MPS for reconstructing stochastic random samples \cite{Caers2004MultiplepointGA,doi:10.1029/2008WR007621,3061447,mosser2017reconstruction}.
Although MPS algorithms achieved in many successful applications \cite{Hermans2015,Huysmans2014, mahmud2015, michael2010, PhysRevE.97.063304, DAVISON201923}, they suffer from some limitations inherent to the simulation algorithms.
These limitations include a computational cost that can be prohibitive for high resolution three dimensional (3D) applications, the presence of visual artifacts in the model realizations, and a low variability between model realizations due to the limited pool of patterns available in a finite-size training image \cite{Emery2014,Li2015}.


Recently, generative models based on deep neural networks have demonstrated remarkable results in terms of image or texture synthesis.
\cite{goodfellow2014deep} introduced the Generative Adversarial Neural Networks (GANs).
GANs are semi-unsupervised deep-learning frameworks for training both a generative model (G model) for capturing the properties of the training sample and a discriminative model (D model) for estimating the probability of a image that comes from the training sample rather than G. 
\cite{MirzaO14cgan} introduced the conditional version of GANs (CGANs), which can be constructed by adding external information (tags or labels) to both training images and the generated images. 
This study showed that it is possible to control the output of generator by conditioning the GANs model on additional information.
CGANs have been applied to various research for image synthesis with different conditional contexts such as categorical image generation, text-to-image synthesis, and semantic manipulation \cite{8411144,yang2018diversitysensitive,DBLP:journals/corr/abs-1711-11585}.  

\cite{RadfordMC15} introduced a class of convolutional neural networks (CNNs) to develop deep convolutional generative adversarial networks (DCGANs) and proposed a set of architectural constraints on CNNs that make DCGANs more stable to train in most of settings.
They applied the architectural constraints to various image samples and demonstrated the applicability of DCGANs in image synthesis.
\cite{mosser2017reconstruction} reconstructed the 3D solid-void structure of porous media by applying a fully convolutional GAN and compared the results with classical stochastic methods. They showed that the fully convolutional nature of the DCGANs can allow an implicit description of the probability distribution represented by 3D image samples, which can be stored and reused to generate multiple realizations of the pore structure very rapidly.

Recently, \cite{kimse2019} successfully applied DCGANs for generating two distinct stochastic properties of drainage networks from Gibb’s models by accounting for directional connectivity information that represents underlying physical processes of the drainage network (i.e., watershed drainage).  
They showed that training samples with physics-informed constraints can significantly improve the statistical properties of complex network connectivity.
\cite{JetchevBV16} proposed spatial GANs (SGANs) utilizing direct and transposed convolutional layers without fully connected layers.
Although SGANs improved computational speed and memory, and produced satisfactory results on certain texture images samples, SGANs are limited to handle a class of textures \cite{BergmannJV17}.
SGANs strongly depend on the size of the training image and also requires deeper convolutional layers to reproduce a larger projective/receptive field for the connectivity of the patterns in TIs than the standard GANs. Hence, SGANs does not make a significant difference in computation efficiency compared to the standard GANs.
Despite many applications on image synthesis using the GANs framework, computational costs are too expensive to generate high-resolution (2D and 3D) images and the image generation is constrained by the size of training image. 

In this study, we propose a spatially assembled GANs (SAGANs) to improve the computational efficiency and the output image  size compared to the standard GANs framework.
The key idea of SAGANs is that the local probability of output images of the generator is estimated by the discriminator, and then assembled into a global probability of the generated images.
The main concepts and advantages of the SAGANs are presented in the next section, followed by the outline of 2D and 3D categorical training image samples and the synthetic experiments with the proposed framework in Section 3.
The results of synthetic 2D and 3D experiments are then compared to demonstrate the performance of the proposed framework in Section 4.
Finally, Section 5 concludes with a summary of the important results of this study, and outlines possible future developments.

\section{Methodology}

In this section we briefly introduce Generative Adversarial Neural Networks (GANs) and Deep Convolutional GANs (DCGANs), and finally describe Spatially Assembled GANs (SAGANs).

\subsection{Generative Adversarial Neural Networks (GANs)}
\label{S:2-1}

\begin{figure}[ht]
    \centering
    \begin{subfigure}[]{1.0\textwidth}
        \includegraphics[width=1.0\textwidth]{ 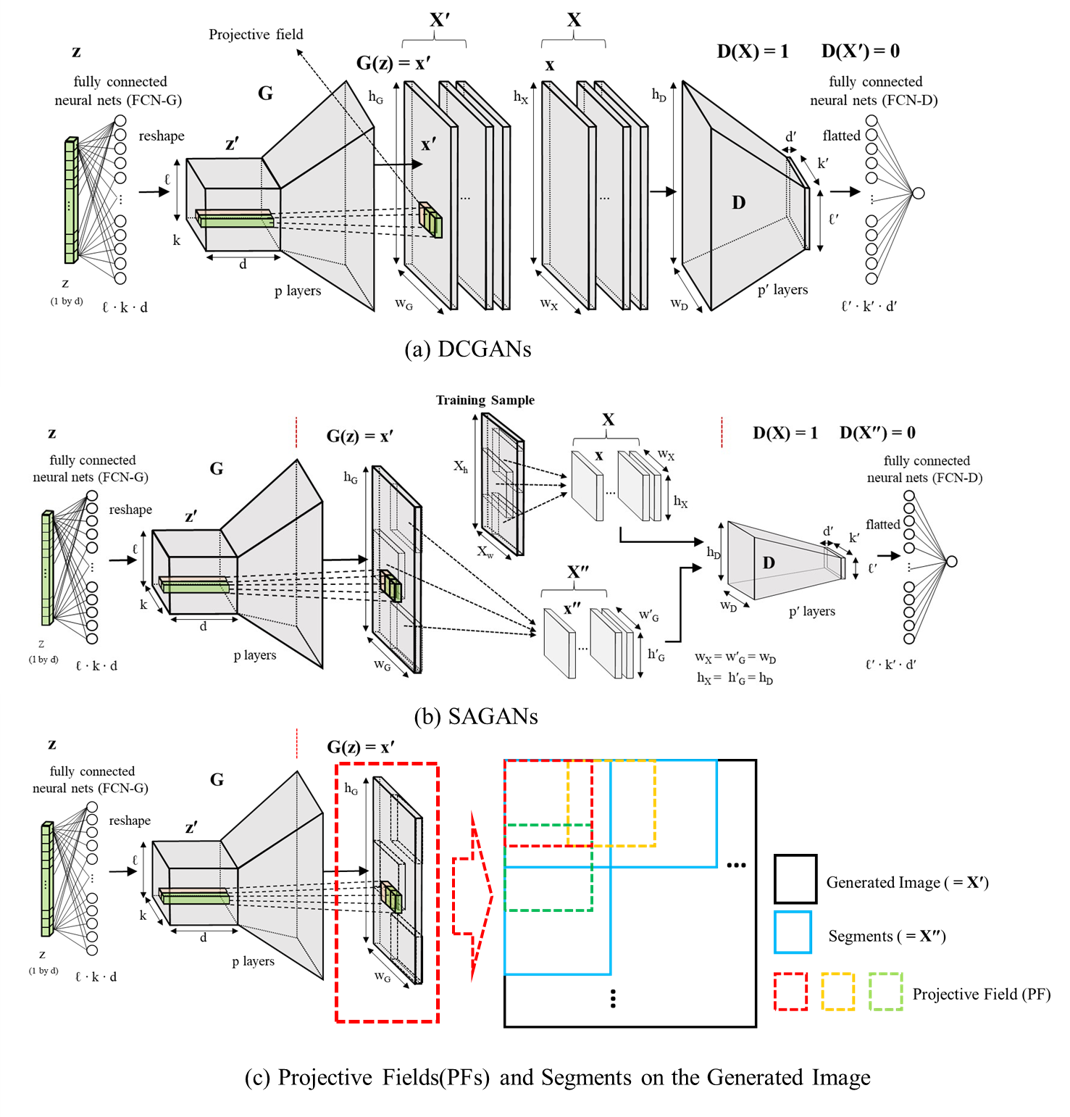}
    \end{subfigure}
    \caption{Frameworks of (a) DCGANs, (b) SAGANs and (c) schematic projective fields and segments in SAGANs}
    \label{fig:f1}
\end{figure}

GANs are a representative deep generative methods to develop generative models via adversarial model \cite{goodfellow2014deep,Goodfellow2016NIPS2T}.
They are defined by an implicit description of the underlying data distribution of training images, which does not rely on an explicit representation of the probability density. 
The generator represents the ability to draw samples from this implicit density.
This is in contrast to explicit density models that try to enumerate the probabilistic generative model of the data e.g., a Markov random field or Markov Chain Product.
GANs train two models simultaneously: a generative model ``$G$'' that captures the ``true'' data generation process from the training sample ($x$), and a discriminative model ``$D$'' that determines whether a sample was from either the generative model ``$G$'' (i.e., false) or the training sample (i.e. true) \cite{mosser2017reconstruction}.
In this way the generator builds a mapping function $G(z;\theta_{g})$ from a prior noise distribution $p_{g}(z)$ to learn the distribution of the ``true'' data $x$ ($p_{g}$), while the discriminator, $D(x;\theta_{d})$, gives a single scalar ($D:\mathbb{R}^{n}\rightarrow [0, 1]$) representing the probability that the discriminator discriminates the ``true'' data $x$ between the ``true'' data and the ``false'' data by $G(z;\theta_{g})$ by $p_{g}$ \cite{goodfellow2014deep,Goodfellow2016NIPS2T}.
 
\begin{equation}
    \min J^{(D)} = -\frac{1}{2}\Big\{\mathrm{E}_{x \sim P_{data}(x)}[\log D(x)]+\mathrm{E}_{z \sim P_{z}(z)}[\log(1- D(G(z)))]\Big\}  
    \label{eq:e1}
\end{equation}

\begin{equation}
    \min J^{(G)} = -\frac{1}{2}\mathrm{E}_{z \sim P_{z}(z)}[\log(D(G(z)))] 
    \label{eq:e2}
\end{equation}
Parameters $(\theta_{d})$ of the $D$ model are adjusted to minimize the mistake ($J^{(D)}$)  in distinguishing the real data (training sample) from the fake created by the generator ({Eq.~\ref{eq:e1}}).
Parameters $(\theta_{g})$ of the G model are adjusted to minimize the log-probability ($J^{(G)}$) of the discriminator being correct for the fake images ({Eq.~\ref{eq:e2}}). 
In other words,  the $G$ model is trained to make $D(G(z)) \rightarrow 1$, while the $D$ model is trained to make $D(G(z))\rightarrow 0$.
Through this procedure, GANs can create the reliable samples by the $G$ model  confirmed by the $D$ model ({Fig.~\ref{fig:f1} (a)}). 

\subsection{Deep Convolutional Generative Adversarial Neural Networks (DCGANs)}
\label{S:2-2}

The DCGANs have been developed to utilize the deep convolutional neural networks (DCNs) in the GANs since the representation of the learned data distribution can be stored in convolutional layers, which is reused to generate samples.
The convolutional nature of DCGANs allows the generation of many samples with the properties similar to the training sample and computational efficiency.
In particular, \cite{RadfordMC15} proposed the architectural guidelines for DCGANs to improve more stable training of GANs by adopting and modifying the DCNs architectures after extensive model exploration.
The proposed guidelines include stride convolutions, no hidden layer in fully connected net, ReLu and LeakyRelu activation function in convolutional layers. The stability of GANs with different training methods can be found in \cite{arora2017generalization,mescheder2018training}

Fig.~\ref{fig:f1} (a) presents the standard (DC)GANs framework. 
The basic concept of the standard DCGANs is that the generator is to map a randomly sampled vector $z$ to a sample $X'$ in the image data space close to the training images $X$, and the discriminator gives a scalar of the probability to indicate if the given image is from the  training images $X$ or from the generated images $X'$ by the generator.
In training process, the standard DCGANs framework imposes a constraint on the dimensions of $X'$ and $X$; the size of $X'$ by the generator should be equal to the size of the training image $X$ since the discriminator is trained with both $X'$ and $X$ to produce the scalar of the probability as in Fig.~\ref{fig:f1} (a). 
In other words, heights and widths of the $G$ outputs, training images and and $D$ model input should be the same, \emph{i.e.}, $h_{G}=h_{X}=h_{D}$, $w_{G}=w_{X}=w_{D}$ as in Fig.~\ref{fig:f1} (a). 
This feature explains that the architecture of the $D$ model is not free from the $G$ model.
Hence, the size of the training images $X$ in the standard DCGANs needs to be the same as the size of the synthetically generated images $X'$. 

In the DCGANs, a randomly sampled vector $z$ given to the $G$ model is reshaped to a set of vector columns $ z'_{(l\times k\times d)}$, where the subscript represents the dimension of the vector columns, through the first hidden fully connected layer. 
Note that vector columns $z'$ are correlated to each other by construction due to the operation of the first hidden fully connected layer on the randomly generated input vector $z$.  
Then, each column of $z'$, $z'_{1 \times 1 \times d}$, generates its piece of field in $X'$, which we call in this work the projective field (PF) \cite{lehky1988network}, through (transposed) convolution layers and defines the local property of the training images.
As these PFs partially overlap each other, all of them are highly correlated with each other to ensure generated images with desired patterns and properties (e.g., connected subsurface channels) once the training is finished.
Consequently the change of a PF affects all pixels of $X'$. 

Hence, any small patches on the image produced by the trained $G$ model will satisfy the local statistical properties learned from the training images for reasonable geomaterial image generation. In the next subsection, we discuss the key idea of the SAGANs framework, which is extended from DCGANs for arbitrary larger size image generation.

\subsection{Spatially Assembled GANs (SAGANs) }
\label{S:2-3}

Our newly proposed method have different implementation details from the standard (DC)GANs.
In the standard GANs, the discriminator $D$ evaluates the probability of an image sampled from the true data generation process, \emph{i.e.}, the training images $X$, or from the generated images $X'$. Thus, the standard GANs are not suitable to creating arbitrary size of images that honors the spatial characteristics of training images. 
On the other hand, SAGANs are designed to evaluate the probability of an image with any arbitrary size originated from the true data generation process based on the subsets (segments) of the training images $X$ and the segments $x''$ of generated image $X'$ as in Fig.~\ref{fig:f1} (b). 
In specific, SAGANs optimizes the generator and the discriminator by investigating the segments of $X'$ and $X$:
\begin{equation}
    G(z) = \{x_{1}^{''},x_{2}^{''}, ... , x_{n}^{''}\}, \;\;\; \textrm{where } X^{'} = \cup^{n}_{i=1} x_{i}^{''},\;\; X \ni \{x_{1},x_{2}, ... , x_{n}\}
\label{eq:e3}
\end{equation}
\begin{equation}    
    \min J^{(D)} = -\frac{1}{2n}\sum_{i=1}^{n}{\{\mathrm{E}_{x \sim P_{data}(x)}[\log D(x_{i})] + \mathrm{E}_{z\sim P_{z}(z)}[log(1- D(x_{i}^{''}))]\}}
\label{eq:e4}
\end{equation}    
\begin{equation}    
    \min J^{(G)} = -\frac{1}{2n}\sum_{i=1}^{n}\mathrm{E}_{z \sim P_{z}(z)}[\log D(x_{i}^{''})] = -\frac{1}{2n}\mathrm{E}_{z \sim P_{z}(z)} [\log(D(x_{1}^{''})D(x_{2}^{''})\cdots D(x_{n}^{''}))]
    \label{eq:e5}
\end{equation}
where $X^{'}$ is a generated image by the generator, $n$ is the number of the segmented images from the generated image and the training image, $x_i^{''}$ is the $i$-th segmented image of $X^{'}$, $x_i$ is the $i$-th segmented image subsampled from the training image $X$, and $\cup$ represents an operation combining the image segments to produce an entire image. 
In SAGANs, each segment is designed to contain multiple PFs to ensure the smoothness and connectivity of the generated whole image. To illustrate this, we present in Fig.~\ref{fig:f1} (c) schematics of $G$ model output with corresponding PFs as red, yellow and green boxes which are the pieces of the generated image corresponding to each column of $z'$. The red PF describes local spatial patterns within one segment while yellow and green PFs are attributed to two adjacent segments. For yellow and green PFs, a column of z’ is trained to generate a part of the whole image that satisfies the desired properties within two segments. As these PFs partially overlap each other, all of them are highly correlated with each other and a change of one PF affects all PFs, \emph{i.e.}, the whole image. 
Since the loss function in SAGANs in Eq.~\ref{eq:e5} is the product of the local probabilities, the local probability from each segment contains the information of multiple PFs and some PFs are divided into multiple segments. Each PF has the local property of the segments where each PF is located. Hence, the product of local probabilities will be minimized in a way that connected information can be learned through the multiple overlapped PFs during training process.

In this way, each local probability of generated segmented images $x''$ is evaluated by the discriminator $D$ and the generator $G$ is optimized to generate seamless images with the local features exhibited from the training data set. This local segment likelihood evaluation approach can be viewed as maximizing one single global likelihood function as in the right hand side of Eq.~\ref{eq:e5}.

This simple but, notable difference enables the generator $G$ to capture the object structure implicitly and synthesize the arbitrary size of images that do not depend on the size of the training samples.
Regardless of the size of images created by the generator, the architecture of the discriminator can be determined depending on the size of the segmented images (i.e., the inputs of a $D$ model). 
Thus, the architecture of the discriminator becomes free from the generator as shown in Fig.~\ref{fig:f1}(b).
This subsequently reduces the computational cost significantly, compared to the standard (DC)GANs.
Note that unlike the standard GANS, SAGANs can create plausible realizations with a few training samples or even a single training sample \cite{tahmasebi2012reconstruction}, which we will demonstrate in the next section.

\section{Development of SAGANs and DCGANs}
\label{S:3}


\subsection{Training Images (TIs)}
\label{S:3-1}
\begin{figure}[ht]
\centering
    \begin{subfigure}{0.8\textwidth}
        \centering
        \includegraphics[width=0.8\textwidth]{ 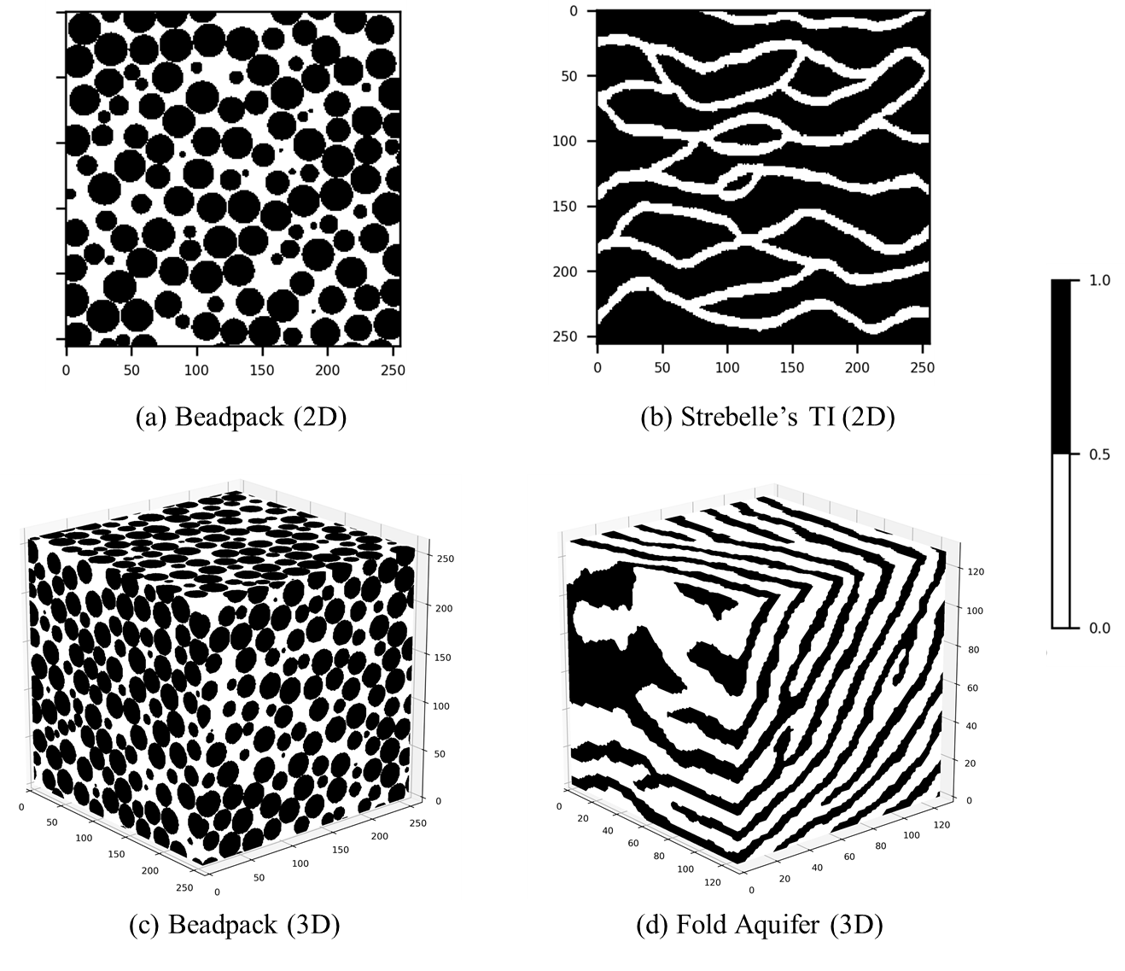}
    \end{subfigure}
\caption[short]{TI samples used in this study: (a) 256 of 2D ($256 \times 256$) Beadpack, (b) one of 2D ($256 \times 256$) Strebelle's TI for 2D realizations, (c) one of 3D ($256 \times 256 \times 256$) Beadpack, (d) one of 3D ($128 \times 128 \times 128$) Fold Aquifer for 3D realizations (complete description of the images can be found in \cite{mariethoz2014multiple}.)}
\label{fig:f2}
\end{figure}

In this study, three TI samples widely used for geostatistical simulation tests were selected to compare the performance of SAGANs with DCGANs for large size image generations and computational efficiency.
Three samples include 2D Strebelle's images \cite{Strebelle2002}, 2D and 3D spherical Beadpack, and 3D categorical Fold Aquifer as shown in Fig.~\ref{fig:f2}.
These samples represent relatively simple geological structures, but have been proven to be very challenging as the training image (TI) for the MPS and GANs \cite{mosser2017reconstruction,Strebelle2002}.
Specifically, Strebelle's TIs and Fold Aquifer consist of continuous channels (or pore space) that are connected in various patterns, while spherical Beadpack patterns are packing of the different size of the spherical particles (beads) which are randomly distributed without being overlapped.
The realization of the discrete nature of Beadpack is also a challenge due to the difficulty of applying GANs for discrete data.
In this study, we focus on how well SAGANs and DCGANs can generate the various long-range connectivity of channels in Strebelle’s TIs and Fold Aquifer, and the spatial distribution and shapes of beads in the spherical Beadpack.
Taking into account the computational time of the experiments and the ease of handling the samples, all TIs are resized into binary 2D squares and 3D cubes with the representation of the pore space or channel (white) and grain solid structure (black).
The TIs in Fig.~\ref{fig:f2} (a) and (b) are used for the 2D realization, and Fig.~\ref{fig:f2} (c) and (d) are used for the 3D realization.

\subsection{Architecture and Training of GANs}
\label{S:3-2}

In this study, we constructed the architecture of deep convolution neural networks for SAGANs and DCGANs based on the guidelines proposed by \cite{RadfordMC15}. \cite{RadfordMC15} identified a family of architectures of deep convolution for stable training across a range of samples and higher resolution and deeper generative models. 
Main architectural features adopted from \cite{RadfordMC15} are the following: (1) Stride convolutions instead of any pooling layers, (2) no hidden layers in a fully connected net (FCN) in both generator and discriminator, (3) ReLU activation in the generator for all layers except for the output that uses \emph{Tanh}, and (4) LeakyReLU activation in the  discriminator for all layers except for the output that uses \emph{sigmoid}. 

Architecture and hyperparameters used in this study are presented in Table \ref{tab:t2}.
\begin{table}[ht]
\centering
\caption{SAGAN architecture and parameters}
\label{tab:t2}
\renewcommand{\arraystretch}{1.2}
\scalebox{0.9}{
\begin{tabular}{ll} 
\hline
\multicolumn{1}{c}{\textbf{Parameter}} & \multicolumn{1}{c}{\textbf{Value}} \\ \hline
Latent Space Dimension, $z$ & 100 \\
Convolutional Layers  / Filters & \begin{tabular}[c]{@{}l@{}}3 or 5 layers, filters ($4/8/16/32/64/128$) \\ Symmetric for D and G model\end{tabular} \\
Kernel Size & $3 \times 3 \times 3$   (for 3D TIs),   $5 \times 5$  (for 2D TIs) \\
Optimizer & Adam with mini-batch \\
Learning Rate / Momentum & $2 \times 10^{-4} / 0.5$ \\
Epoch / Mini-batch Size & Max 100,000 / 4 to 64 \\
\begin{tabular}[c]{@{}l@{}}Dropout rate / \\ Batch normalization Momentum\end{tabular} & 0.25 / 0.8 \\
Activation Function & \begin{tabular}[c]{@{}l@{}}ReLu, tanh (G model) / \\ LeakyReLu with alpha  0.2, sigmoid (D model)\end{tabular} \\
Loss Function & Binary Cross-entropy \\ \hline 
\end{tabular}%
}
\end{table}

The $D$ model is composed of two convolution layers with three or five of kernel size, and the output was converted to the probability (0 (false) to 1 (true)) using the \emph{sigmoid} activation function.
In the $G$ model, the noise $z$ with a dimension of 100 was drawn from a Gaussian distribution and given to the FCN.
The FCN reshaped $z$ to a set of tensor ($z'$) before passed to the convolutional layers.
The deconvolution layers with the same kernel size in the $D$ model was used, and the output is converted by a stride convolution layer into the same size image as the given training images or segments. 
Both $D$ and $G$ models were applied and trained by the Adam (adaptive momentum estimation) optimization algorithm with a learning rate of 0.0002 with the momentum ($\beta_{1}$) of 0.5, and a total of 100,000 epochs.
Dropout with a probability of 0.25 was applied to the convolution layers of the $D$ model.
The loss of GANs was estimated by binary cross-entropy function (Eqs.~\ref{eq:e1} and ~\ref{eq:e4}).
The different size (pixel and voxel resolution) of images from the TIs are synthesized by DCGANs and SAGANs.
Depending on the synthesized size, 3 to 5 depth of convolutional layers and 4 to 128 channels (filters) per a convolutional layer were applied.
A larger number of epochs with a small learning rate and dropout on the $D$ model and batch normalization were applied for the stabilized training and the convergence of both $G$ and $D$ models that have the symmetric architecture.
 
All computational works in this study were performed using the same computer equipped with two NVIDIA TITAN-V GPU cards (5,120 cores / 12G GPU memory) and Intel i9-9900X CPU (10 cores) with 64G RAM.


\subsection{Evaluation of the Synthesis Results}
\label{S:3-3}

Two-point probability and two-point cluster functions were used to evaluate the texture image synthetic performance of both SAGANs and DCGANs in this study.
The two-point probability function ($S_{2}$) estimates the probability of how  two points (pixels for 2D and voxels for 3D digitized images) are related to each other, which is often called auto-correlation.
$S_{2}$ is used for structural information about the degree of spatial dependence of a random field.

\begin{equation}
    S_{2}(r) = P\{x \in \Phi ; x + r  \in \Phi\} \;\; for\;\; x, r \in  \mathbb{R}^{n}
\end{equation}
where $P$ is the probability that two points $x$ and $x+r$ are separated by the lag vector $r$ in the same phase of interest $\Phi$.
If the two points coincide $(r = 0$ in the phase $\Phi)$, the probability is the same as the probability that one point is the phase $\Phi$, which is simply equal to the (volume) fraction $\varphi$ of the phase $\Phi$,

\begin{equation}
    \varphi = \frac{V_{\phi}}{V_{total}}, \quad \quad \quad  S_{2}(r = 0) = \varphi \\
\end{equation}
When the two points are completely uncorrelated and far apart, each point has a probability $\varphi$ to be the phase $\Phi$,

\begin{equation}
    S_{2}(r \rightarrow \infty) = \varphi^{2}
\end{equation}
$S_{2}$ decays from $\varphi$ towards the asymptotic value of $\varphi^{2}$.

Two-point cluster function $C_{2}$ estimates the probability that two points are in the same cluster of the phase of interest when the image sample consists of some isolated clusters. 
$C_{2}$ is used for representing the statistically homogeneous distributions of identical inclusions \cite{torquato2013random}. 
This connectivity function is defined as follows,

\begin{equation}
C_{2}(r) = P\{\mathcal{C}(x) = \mathcal{C}(x+r)\} \quad for \quad x \in \Phi_{i}
\end{equation}
where $\mathcal{C}$ is an indicator of the cluster in $\Phi_{i}$.

\section{Results and Discussion}
\subsection{2D \& 3D Earth Texture Image Synthesis}
\label{S:4-1}

\begin{figure}[h]
    \centering
    \begin{subfigure}[]{0.8\textwidth}
        \includegraphics[width=\textwidth]{ 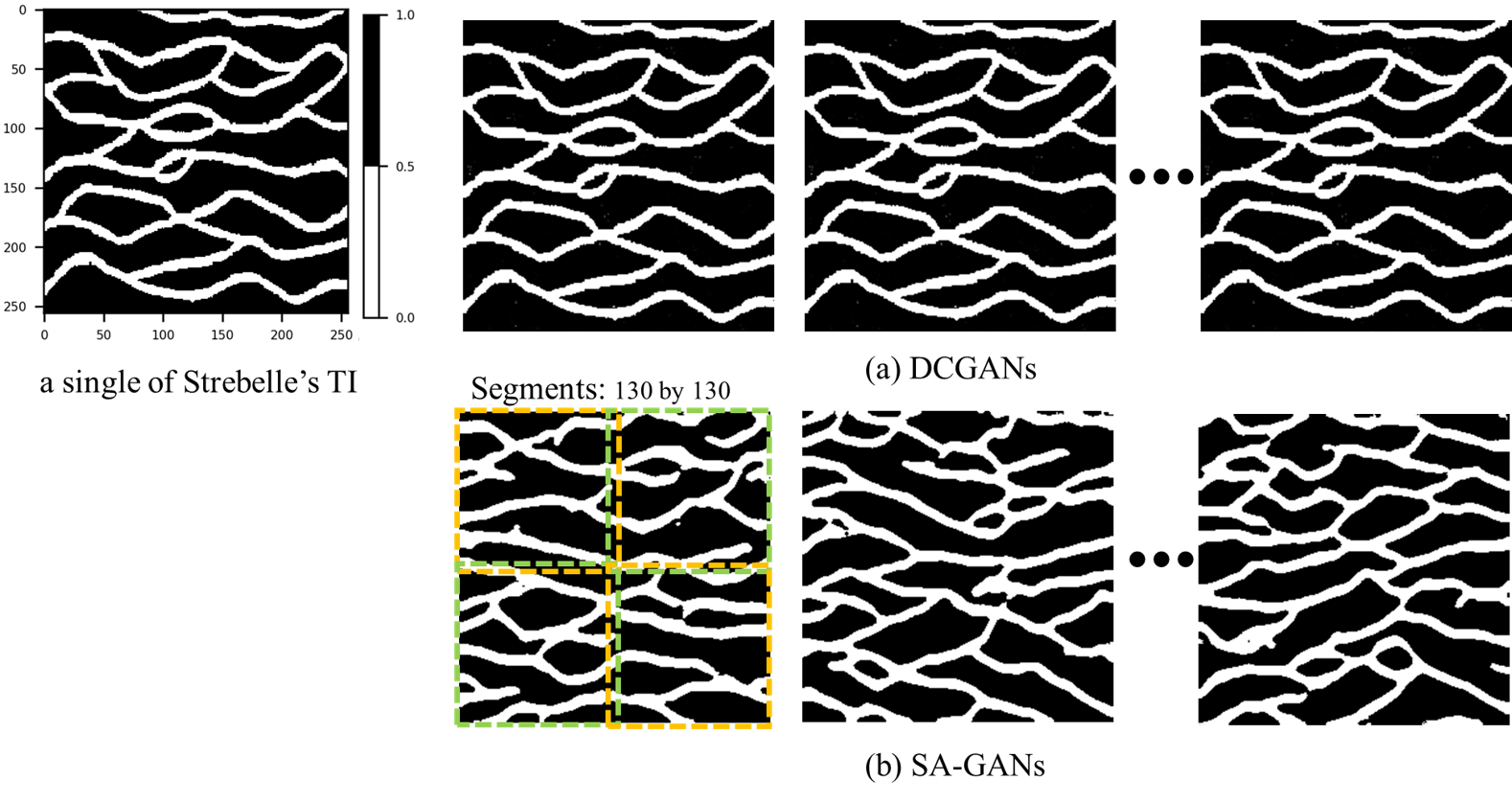}
        \setlength{\belowcaptionskip}{-8pt}
    \end{subfigure}

    \caption[short]{Comparison of the synthesis results of DCGANs and SAGANs on a single of Strebelle's TI ($256 \times 256$ pixels)} 
    \label{fig:f3}
\end{figure}

In this section we compared the results obtained from SAGANs with those from DCGANs using the 2D $\&$ 3D TIs presented in Fig.~\ref{fig:f2}.
First, we consider a simple example using the Strebelle's TI to illustrate the effectiveness of the proposed method. The single Strebelle's TI consists of connected channels (white color) representing the long-range connectivity and pattern diversity. Several images generated from DCGANs and SAGANs are plotted as in Fig.~\ref{fig:f3} 
Notably the DCGANs produced the same realization of images as the TI, while the SAGANs produced different realizations with various patterns visually similar to those in the Strebelle's TI. 
This is mainly due to the architecture of DCGANs and the lack of diverse training data; the $D$ model in the DCGANs determines whether the image created by the $G$ model is the same as the TI (true) or not (fake). Then, the $G$ model in the DCGANs is trained to generate exactly the same image as the given single TI, resulting in the same realization as the TI (Fig.~\ref{fig:f3} (a)). Since in many subsurface data survey only a limited number of geomaterial images would be available \cite{yoon2013nanopore}, direct application of DCGANs to geostatistical simulation would not be satisfactory. 
\begin{figure}[h]
    \centering
    \begin{subfigure}[]{0.8\textwidth}
            \includegraphics[width=\textwidth]{ 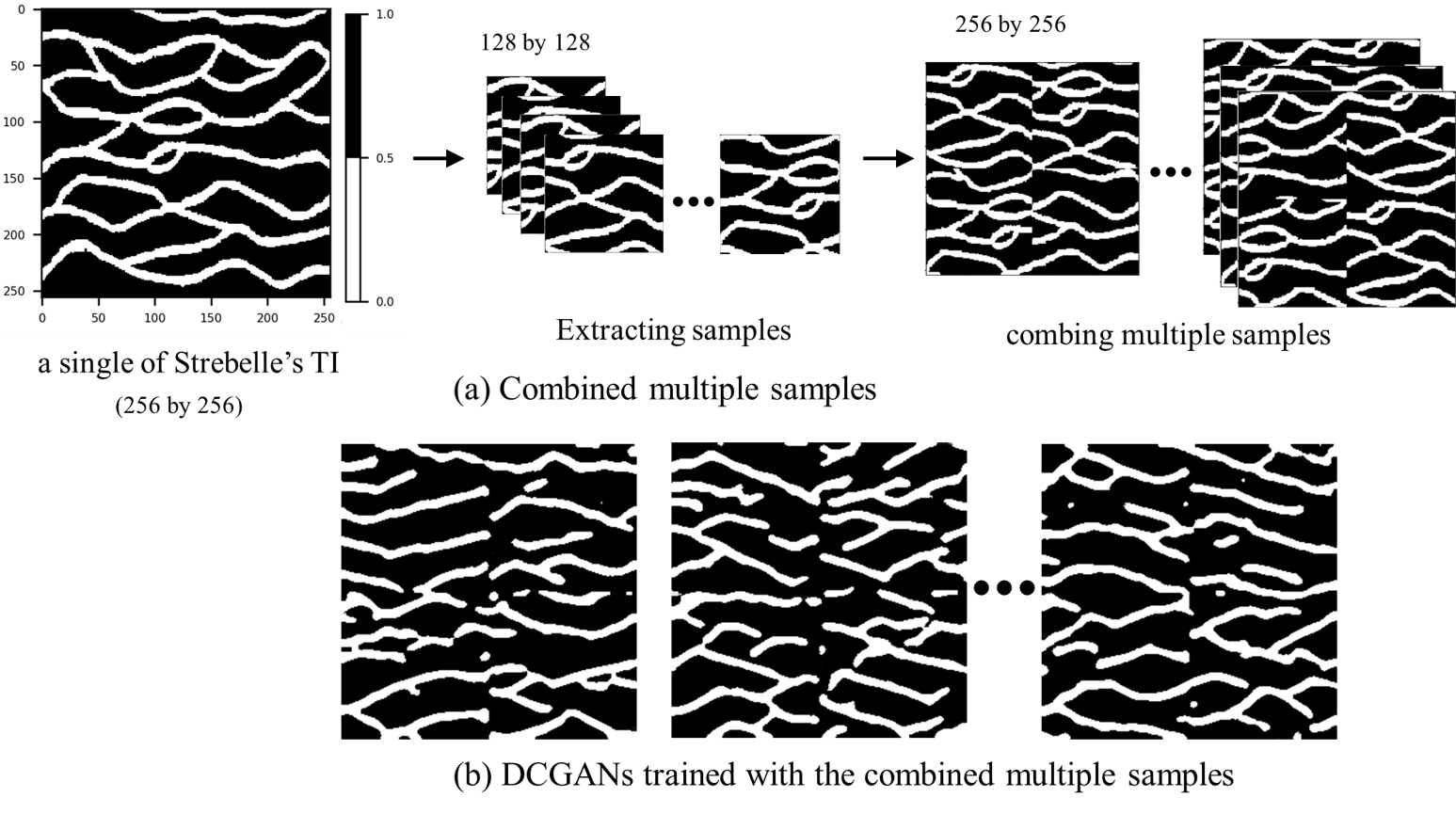}
            \setlength{\belowcaptionskip}{-8pt}
    \end{subfigure}
    
    \caption{Synthesis results of DCGANs trained with the combined multiple samples extracting from a single of Strebelle's TI }
    \label{fig:f4}
\end{figure}
\begin{figure}[ht]
\centering
    \begin{subfigure}{0.85\textwidth}
        \centering
        \includegraphics[width=0.8\textwidth]{ 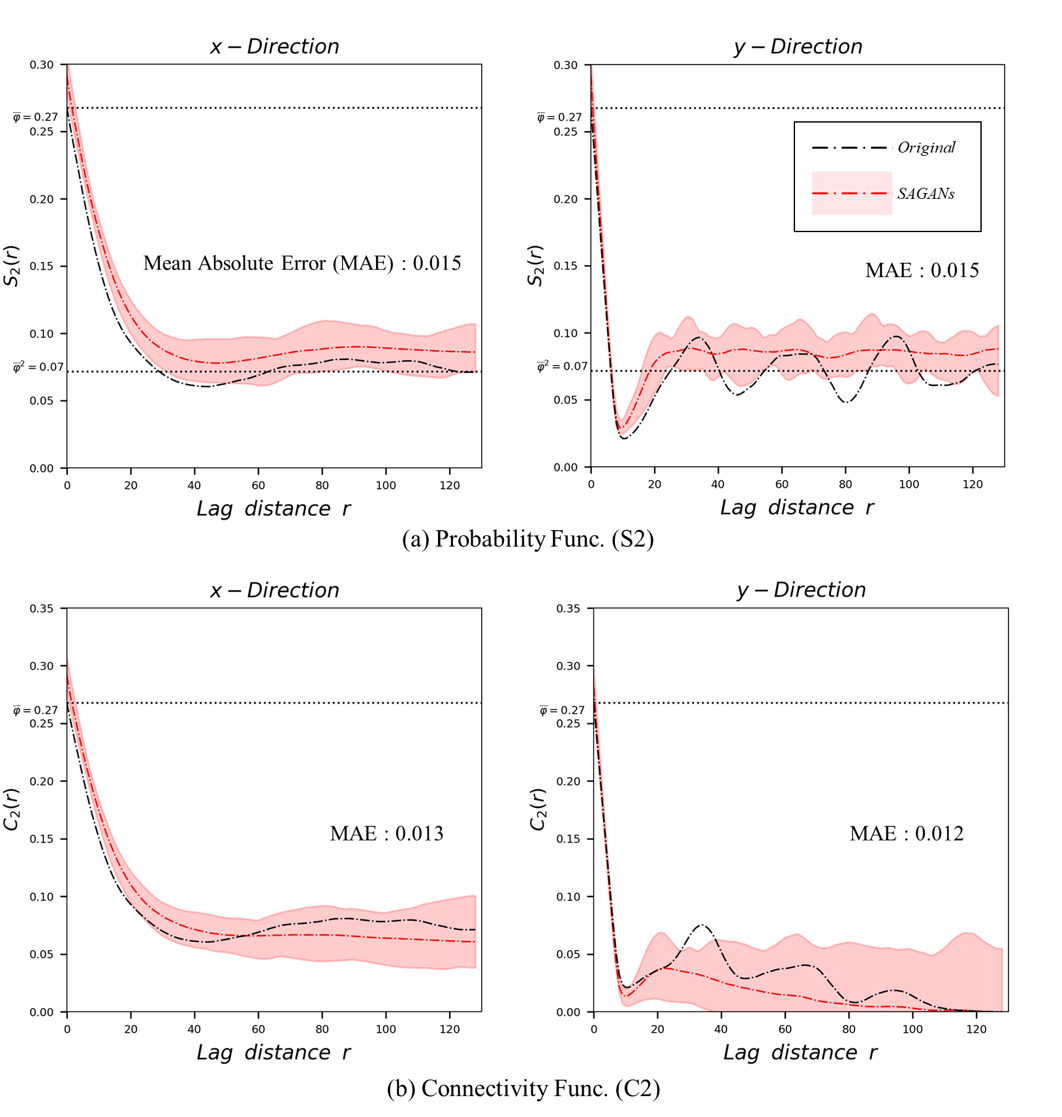}
        \setlength{\belowcaptionskip}{-10pt}
    \end{subfigure}
\caption[short]{Comparison of $S_{2}$ and $C_{2}$ on the realizations and Strebelle's TI}
\label{fig:f5}
\end{figure}

In contrast to the DCGANs, the image generated by the $G$ model in the SAGANs is divided into several segments, and then each segment is compared with the same size of segments extracted from a single TI (Fig.~\ref{fig:f1} (b)).
In this example, both generated image and the TI were divided into four segments as in Fig.~\ref{fig:f3} (b).
Each segment has a size of 130 by 130 with 4 pixels overlapped vertically and horizontally to account for the long-range connectivity. 
The arrangement of the segments from the TI and the structures in place build a pool of patterns available.
The architecture of the SAGANs can capture the object structure implicitly from the segments through hierachical assembly between PF and segments, hence the SAGANs can synthesize the various pattern of textures with statistical properties (e.g., the long-range connectivity) even using a single TI. Although DCGANs can synthesize the various textures using the multiple segments extracted from a single image as TI like the way SAGANs extract information from the TI (Fig.~\ref{fig:f4} (a)), without any architecture or training image modifications, DCGANs cannot generate the smoothly connected channels properly since the seam between the combined multiple segments in the training images is also trained as a pattern (Fig.~\ref{fig:f4} (b)). 

\begin{figure}[ht]
\centering
    \begin{subfigure}{0.8\textwidth}
        \centering
        \includegraphics[width=0.8\textwidth]{ 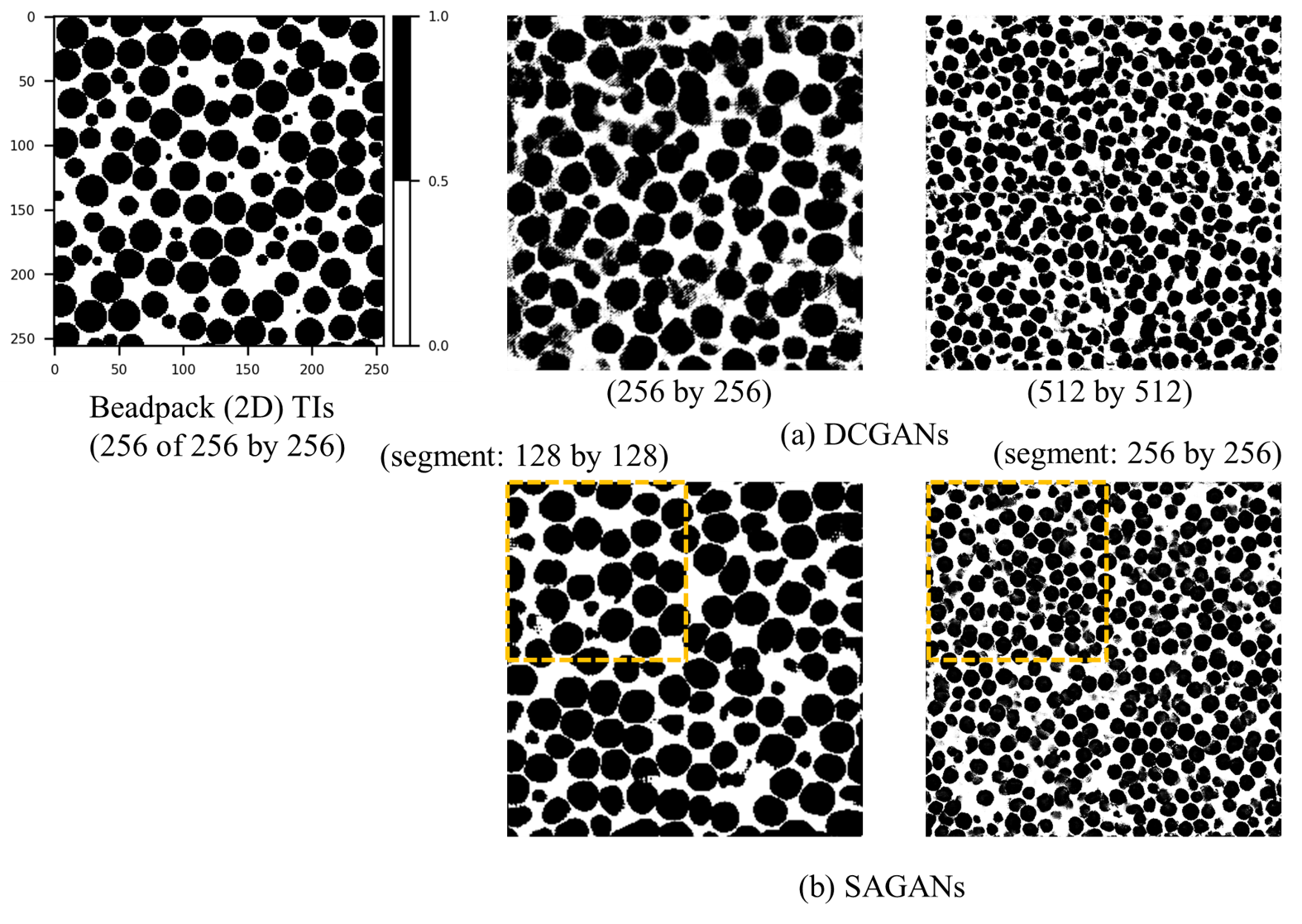}
        \setlength{\belowcaptionskip}{-30pt}
    \end{subfigure}
\caption[short]{Comparison of the synthesis results of DCGANs and SAGANs on Beadpack TIs}
\label{fig:f6}
\end{figure}
\begin{figure}[!h]
\centering
    \begin{subfigure}{0.8\textwidth}
        \centering
        \includegraphics[width=0.8\textwidth]{ 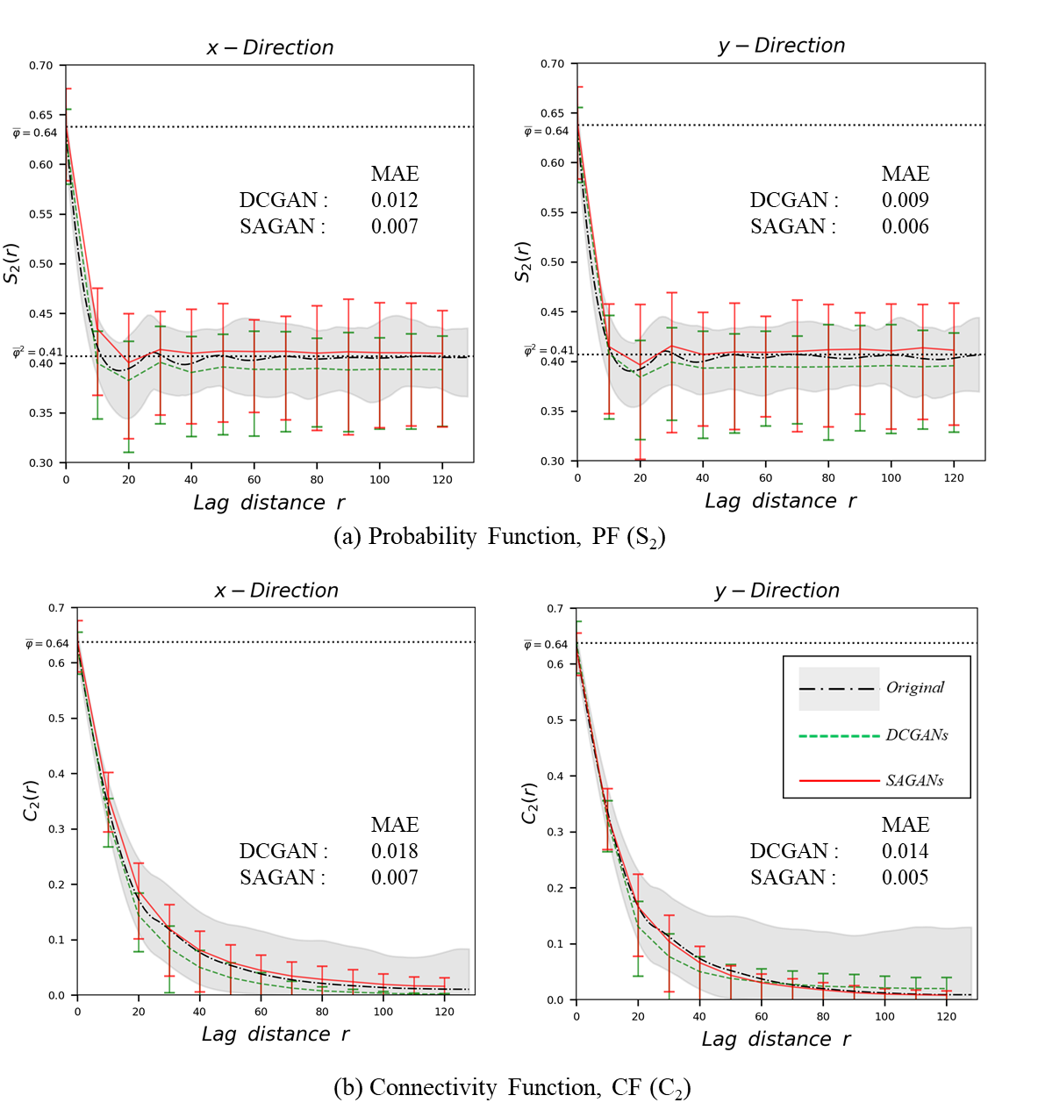}
        \setlength{\belowcaptionskip}{-10pt}
    \end{subfigure}
\caption[short]{Comparison of $S_{2}$ and $C_{2}$ over the realizations and spherical Beadpack TIs; the green error bar represents the range of $S_2$ and $C_2$ values of spherical beadpack images generated by DCGANs with respect to the lagged distance $r$. The red error bar represents the range of $S_2$ and $C_2$ values of spherical beadpack images generated by SAGAN w.r.t a lagged distance $r$. The grey intervals represent the range of $S_2$ and $C_2$ values of training (original) images w.r.t a lagged distance $r$.}
\label{fig:f7}
\end{figure}

To compare the spatial similarities between the generated image and the TI, two directional metrics ($S_{2}$ and $C_{2}$) for the pore phase (channels in white) of twenty generated images were calculated as shown in Fig.~\ref{fig:f5}.
Note that $S_{2}$ and $C_{2}$ of the realizations by DCGANs are identical to those of the TI (`Original' line in Fig.~\ref{fig:f5}) since the same realizations as a single TI were always produced.
On the other hand, the realizations by SAGANs show a range of fluctuation in $S_{2}$ and $C_{2}$ (a shade in red in Fig.~\ref{fig:f5}) but show similar average values (red line) to those of from the TI.
They also have the similar volume fraction $\varphi$ (porosity) of 0.29 compared to 0.27 of the TI.
This result indicates that SAGANs can generate statistically similar images with various patterns even using a single TI.

\begin{figure}[h]
\centering
    \begin{subfigure}{0.8\textwidth}
        \centering
        \includegraphics[width=1.0\textwidth]{ 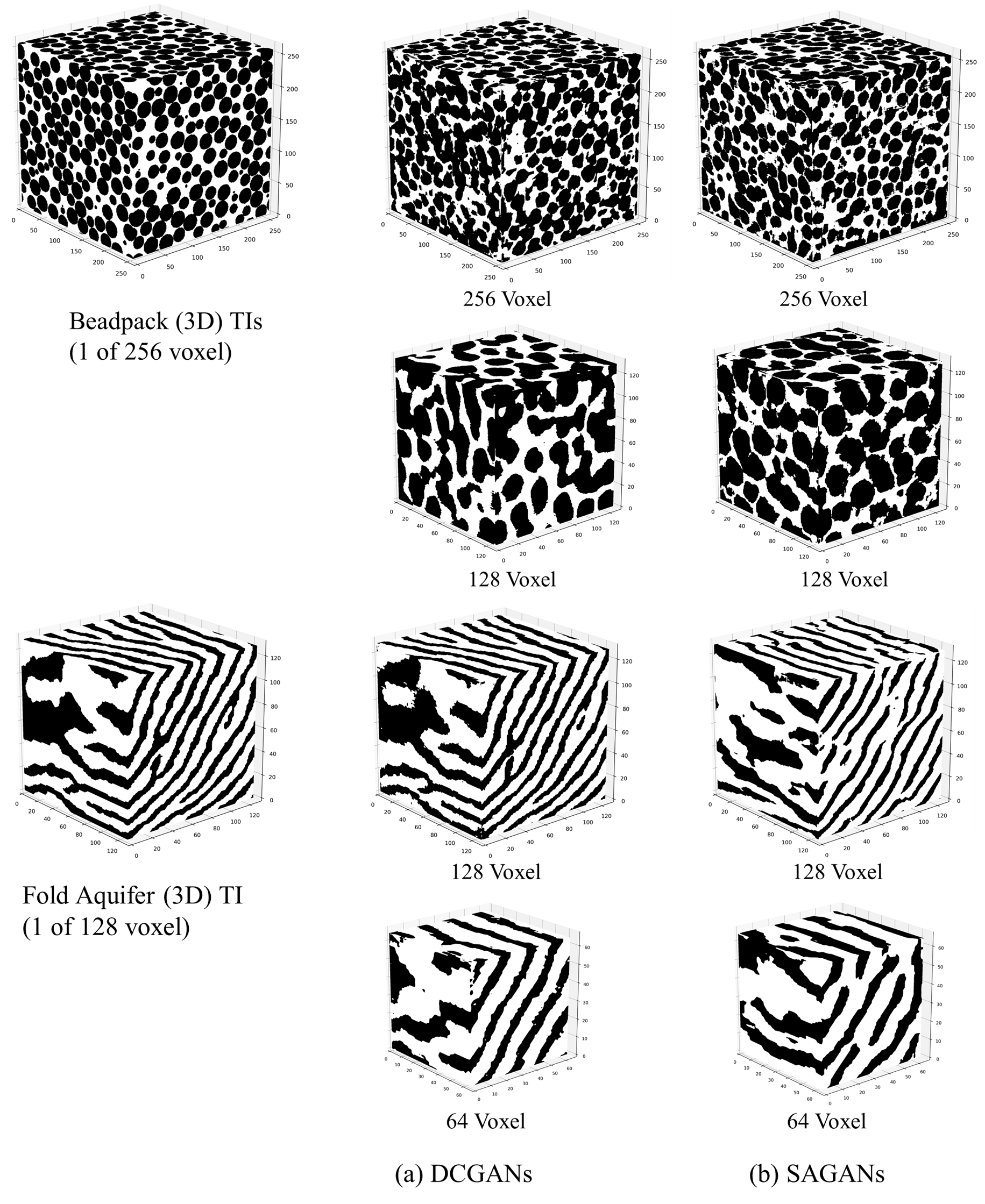}
        \setlength{\belowcaptionskip}{-30pt}
    \end{subfigure}
\caption[short]{Comparison of the 3D synthesis results of DCGANs and SAGANs on the spherical Beadpack TIs (3D)}
\label{fig:f8}
\end{figure}

Our second 2D realization uses a total of spherical Beadpack TIs  (Fig.~\ref{fig:f2} (a)) for the texture synthesis on the spatial distribution and shapes of the particles (structural properties). Fig.~\ref{fig:f6} shows some examples of the realizations on the the spherical Beadpack TIs.
Four segments with a quarter size of the generated image and the TIs were used in SAGANs.
Beads in the realizations by SAGANs have more spherical shape and less overlapped with each other (well-spread) than those by DCGANs (Fig.\ref{fig:f6} (b)).
Both DCGANs and SAGANs achieve a small error in the averaged values of $S_{2}$ and $C_{2}$ for each twenty generated images (Fig.~\ref{fig:f7}), with the SAGANs slightly better the DCGANs.
In contrast to the first case using a \emph{single} TI, the second case demonstrates that with a large number of TIs both DCGANs and SAGANs are able to the images with the basic statistics of the TIs.

\begin{figure}[!h]
\centering
    \begin{subfigure}{0.8\textwidth}
        \centering
        \includegraphics[width=.9\textwidth]{ 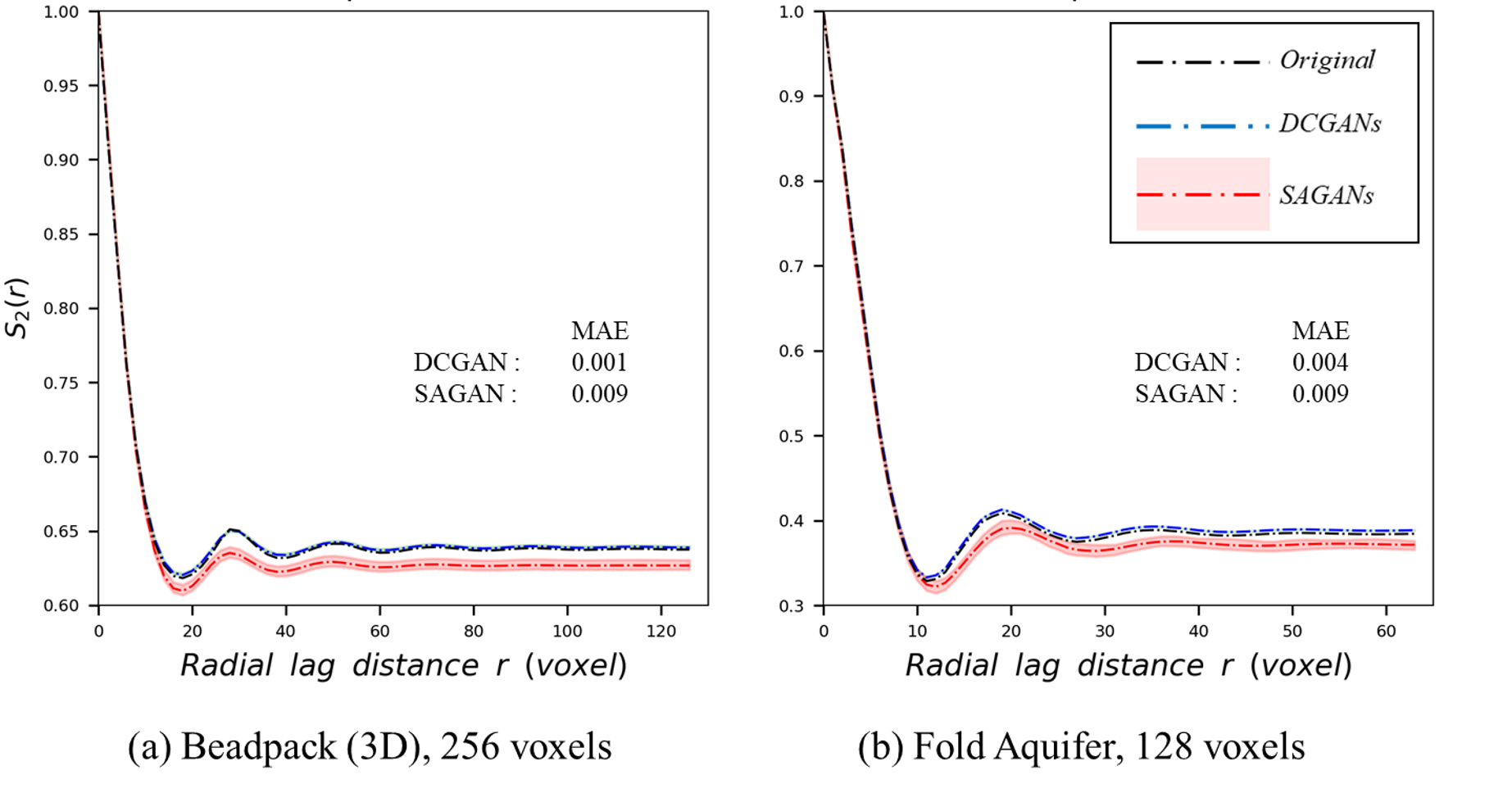}
    \end{subfigure}
\caption[short]{Normalized isotropic $S_{2}$ comparison of the 3D synthesis results by DCGANs and SAGANs}
\label{fig:f9}
\end{figure}

In the 3D case, we used spherical Beadpack TIs and Fold Aquifer TIs (Fig.\ref{fig:f2} (c), (d)).
Similar to the 2D cases, we evaluated the performance of SAGANs and DCAGNs on the 3D realization of the long-range connectivity and the spatial distribution of the particles.
Fig.~\ref{fig:f8} shows examples of the 3D realizations by both DCGANs and SAGANs. 
To compare the similarity of the 3D realizations, the $S_{2}$ over the radial lag distance was calculated and normalized for the randomly selected 20 realizations.
The $C_{2}$ values are not reported since they are very identical to the $S_{2}$ statistics.
Comparison of the normalized isotropic $S_{2}$ between the realizations and TIs are shown in Fig.\ref{fig:f9}. 
Since the single 3D TI was used, the DCGANs generate the same realization as the TI, hence the $S_{2}$ of TIs is identical to one of the DCGANs.
As in the first 2D realization case, the $S_{2}$ of the SAGANs shows a relatively small error compared to the TIs and a small variation, indicating that the SAGANs can produce the 3D realizations with diversity even using the single 3D TI.
\subsection{Scalability and Computational Efficiency}
\label{S:4-3}

\begin{figure}[h]
    \centering
    \begin{subfigure}[]{0.8\textwidth}
        \includegraphics[width=\textwidth]{ 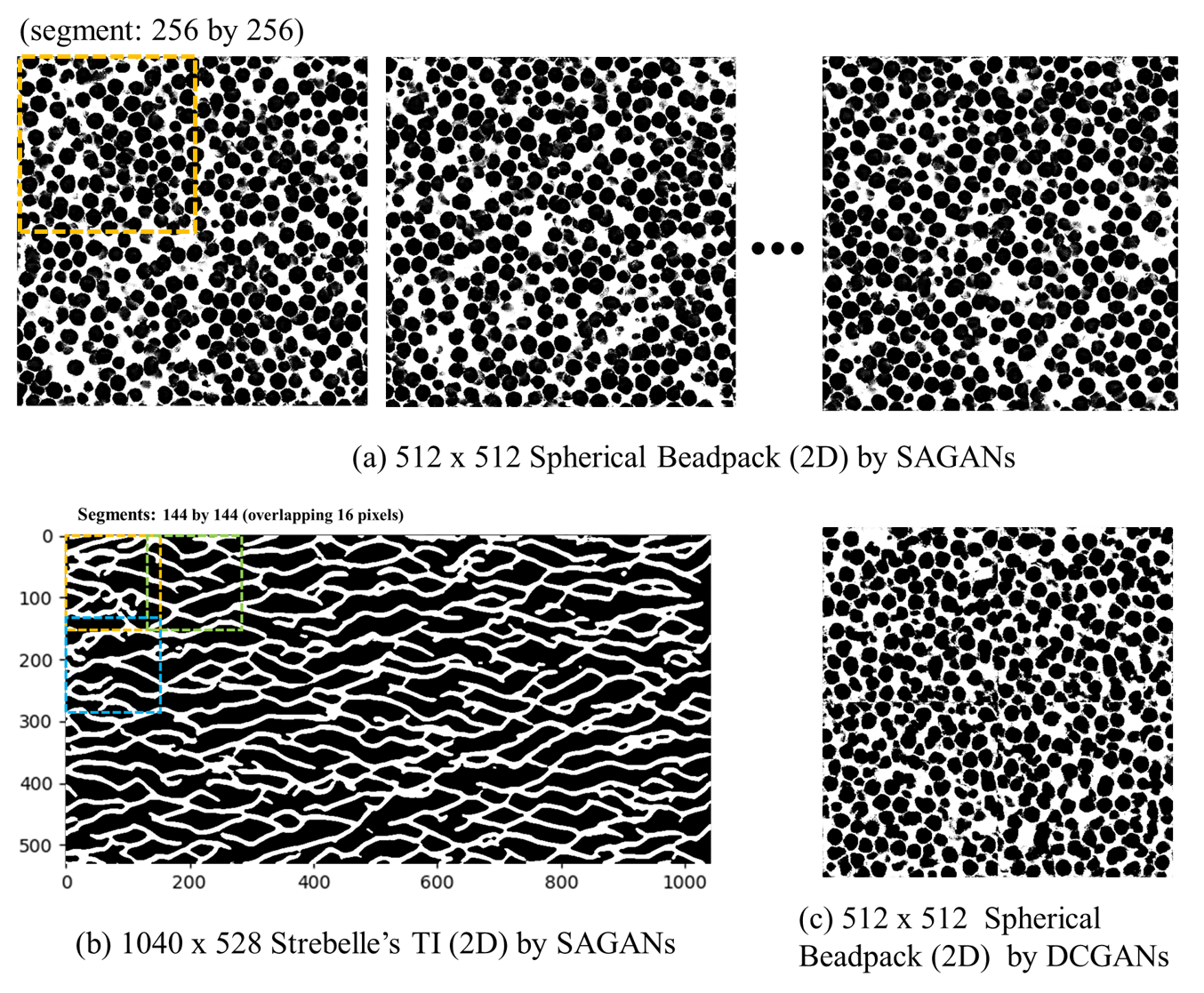}
    \end{subfigure}
    \caption{Examples of the large size 2D realization by SAGANs and DCGANs}
    \label{fig:f10}
\end{figure}

In this section, we focus on the scalability in terms of the output texture size and computational efficiency of the SAGANs.
As explained previously, the size of model output is not limited by the TI size in the SAGANs, hence, we tested this conceptual advance with a smaller size of the images extracted from the TIs to generate a large size of image.
Fig.~\ref{fig:f10} (a-b) shows examples of the arbitrary large size of 2D realization by the SAGANs.
In the realization of larger size of 2D spherical Beadpack ($512 \times 512$ pixels) by SAGANs, the created image of $512 \times 512$ pixels were divided into four segments of $256 \times 256$ pixels, and then the probability of each segment was evaluated with the segments of $256 \times 256$ pixels from Beadpack TIs. 
For the realization of larger size of Strebelle's TI-based channels ($1,040 \times 528$ pixels), 32 segments of $144 \times 144$ pixels divided with overlapping 16 pixels in the created image $1,040 \times 528$ pixels were evaluated with the segments of $144 \times 144$ pixels from $256 \times 256$ pixels of Strebelle's TI.

In the Fig.~\ref{fig:f10} (a-b), the spherical beads and the long-range connectivity can be verified visually.
The larger size images can be generated by DCGANs in the same way shown as in Fig.~\ref{fig:f4} (a). 
However, the realizations show the seam between the combined original TIs as the realizations in Fig.~\ref{fig:f4} (b) and Fig.~\ref{fig:f10} (c).
In addition, building the large size of TIs from the small size TIs requires considerable time and computational resources for reading and allocating the data in the memory to train the GANs, especially in building 3D TIs.
Hence, the DCGANs have the limits in generating the arbitrary large size of images due to the cropping area (or the seam) created by connecting multiple images from the original TIs. 
On the other hand, SAGANs produce the arbitrary large seamless realizations with simply changing the dimension of $z'$ in the generator (G) even in a sing TI. 

\begin{figure}[h]
    \centering
    \begin{subfigure}[]{0.8\textwidth}
        \includegraphics[width=0.8\textwidth]{ 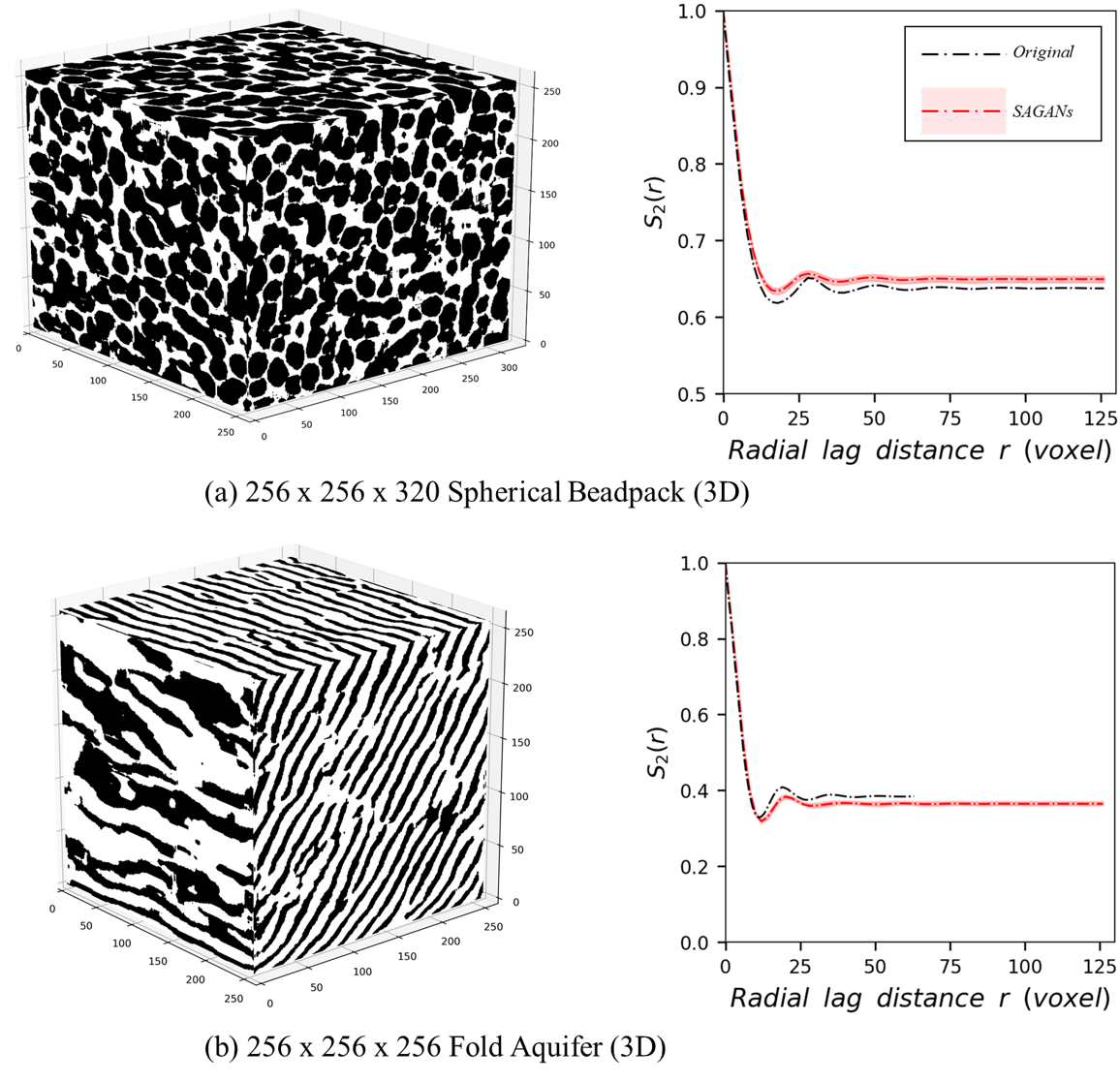}
    \end{subfigure}
    \caption{Examples of the arbitrary large size 3D realizations by SAGANs}
    \label{fig:f11}
\end{figure}

The size and positions of segments also can be arbitrarily adjusted according to the size of the image to be generated for the characteristics of the original TIs.
Fig.~\ref{fig:f11} shows examples of the arbitrary large size of 3D realizations from a single of TI by SAGANs.
In the realization of $256 \times 256 \times 320$ size spherical Beadpack (3D), segments of the 64 voxels were applied.
For the realization of $256 \times 256 \times 256$ size Fold Aquifer TI (3D), segments of the 64 voxels overlapping by 3 voxels were applied considering the long-range connection.
The $S_{2}$ graph on the twenty realizations by SAGANs shows a small error to the graph on TIs and with some variations.
This indicates that SAGANs could produce the 3D realizations of the arbitrary larger size and with diversity even in a single of 3D TI.
\begin{table}[ht]
\centering
\caption{Comparison of training time between SAGANs and DCGANs}
\label{tab:t3}
\renewcommand{\arraystretch}{1.2}
\scalebox{0.9}{
\begin{tabular}{ccccl}
\hline
\textbf{TIs} & \textbf{Generated Size} & \textbf{Model} & \textbf{Segment size} & \multicolumn{1}{c}{\textbf{\begin{tabular}[c]{@{}c@{}}Training Time (Sec.)\\ (Decrease)\end{tabular}}} \\ \hline
\textbf{Beadpack (2D)} & $256 \times 256$ & DCGANs &  & 45.54 \\
(256 of $256 \times 256$) &  & SAGANs & $128 \times 128$ & 27.70 (-39$\%$) \\ \cline{2-5} 
\multicolumn{1}{l}{} & \multirow{2}{*}{\begin{tabular}[c]{@{}c@{}}$512 \times 512$\\ (Larger than TIs)\end{tabular}} & DCGANs &  & 120.92 \\
\multicolumn{1}{l}{} &  & SAGANs & $128 \times 128$ & 51.26 (-58$\%$) \\ \hline
\textbf{Strebelle TI (2D)} & $256 \times 256$ & DCGANs &  & 53.70 \\
(1 of $256 \times 256$) &  & SAGANs & $130 \times 130$ & 22.92 (-57$\%$) \\ \cline{2-5} 
\multicolumn{1}{l}{} & \multirow{2}{*}{\begin{tabular}[c]{@{}c@{}}$512 \times 512$\\ (Larger than TIs)\end{tabular}} & DCGANs &  & 196.94 \\
\multicolumn{1}{l}{} &  & SAGANs & $131 \times 131$ & 61.17 (-69$\%$) \\ \cline{2-5} 
\multicolumn{1}{l}{} & \begin{tabular}[c]{@{}c@{}}$528 \times 1040$\\ (Larger than TIs)\end{tabular} & SAGANs & $144 \times 144$ & 120.86 \\ \hline
\textbf{Beadpack (3D)} & $128 \times 128 \times 128$ & DCGANs &  & 909.83 \\
(1 of $256 \times 256 \times 256$) &  & SAGANs & $64 \times 64 \times 64$ & 250.78 (-72$\%$) \\ \cline{2-5} 
 & $256 \times 256 \times 256$ & DCGANs &  & 6531.66 \\
 &  & SAGANs & $64 \times 64 \times 64$ & 839.30 (-87$\%$) \\ \cline{2-5} 
 & \begin{tabular}[c]{@{}c@{}}$256 \times 256 \times 320$\\ (Larger than TIs)\end{tabular} & SAGANs & $64 \times 64 \times 64$ & 1013.89 \\ \hline
\textbf{Fold Aquifer (3D)} & $64 \times 64 \times 64$ & DCGANs &  & 231.00 \\
(1 of $128 \times 128 \times 128$) &  & SAGANs & $34 \times 34 \times 34$ & 81.30 (-65$\%$) \\ \cline{2-5} 
 & $128 \times 128 \times 128$ & DCGANs &  & 1066.1 \\
 &  & SAGANs & $66 \times 66 \times 66$ & 234.06 (-78$\%$) \\ \cline{2-5} 
\multicolumn{1}{l}{} & \begin{tabular}[c]{@{}c@{}}$256 \times 256 \times 256$\\ (Larger than TIs)\end{tabular} & SAGANs & $67 \times 67 \times 67$ & 839.26 \\ \hline
\end{tabular}
}
\end{table}

DCGANs could not generate larger images than the original TI size (larger than 256 voxels of Beadpack, larger than 128 voxels of Fold Aquifer) due to the lack of TIs and GPU memory limitations. 
As mentioned above, the size of the generated output in the G model should be identical to the size of the input of D model in the standard GANs (DCGANs) framework.
Therefore, the larger the output is generated, the larger architecture of D model is required at the same time, which increases the computational load.
The D model runs not only when the D model is trained, but when the G model is trained.
Consequently, this architectural constraint increases both the computational load and computational time in training the GANs significantly.
However, in the SAGANs the architecture of the D model is free from the G model. 
The size of the inputs of a D model can be set relatively small regardless of the size of the output generated by the G model.
Therefore, SAGANs can produce the larger size realizations with the low computational cost.
Table.~\ref{tab:t3} shows the training time per 1,000 epoch.
In each case, the symmetrical architecture for G and D model was applied with the same parameters (Table.~\ref{tab:t2}). 
The difference between DCGANs and SAGANs in each case is the size of the inputs (the size of segment) given to the D model.
The results in Table.~\ref{tab:t3} show that the training time of the SAGANs is less than the training time of the DCGANs, and the decrease is larger as the size of generated outputs increases.
Fig.~\ref{fig:f11} and Table.~\ref{tab:t3} indicate that the texture synthetics with the SAGANs significantly improve the increase in the image size of realizations and the computational time for high-dimensional applications compared to the standard GANs framework. The detailed list of DCGAN and SAGAN architecture and hyperparameters of each test can be found in the Appendix. 

\section{Conclusions}
This study proposed the new GANs framework, spatially assembled GANs (SAGANs) that can generate output images of an arbitrary large size regardless of the size of training images with computational efficiency.
The SAGANs have no architectural constraint, in contrary to the standard GANs (DCGANs) framework where the size of the generated output image in the G model should be identical to the size of the input image of D model.
This enables the SAGANs to produce the realizations with the scalability in terms of the output size and computational efficiency.

The performance of the SAGANs for generating the images was evaluated using widely-used 2D and 3D earth texture image samples with two-point probability and two-point cluster function.
Both SAGANs and DCGANs generated the texture images close to the basic statistics and morphological characteristics of the TIs such as porosity, long-range connectivity, the spatial distribution and shapes.
However, the DGGANs could not produce various patterns of the realizations when a single of TI is given, and could not produce the arbitrary large size of realizations due to the cropping area (or the seam) created by combining the multiple original TIs.
In addition, the DCGANs could not produce large size of 3D realization due to the lack of computational memory.
However, the SAGANs improve these limitations to produce large seamless realizations with a variety of patterns, even using a single TI.
The SAGANs also produced the larger size of 3D realizations with a low computational time and load so that the applications for high-dimensional problems are now achievable.

The SAGANs can be applied in various ways depending on how to adjust the size and positions of segments. 
For example, if a specific field in generated images by the G model is evaluated by the same segment from TIs, the part of all the generated images has the same property as in the specific segment.
In this way, a specific part of the texture image can be synthesized.
If various segments from TIs with quite different features are used, the SAGANs can produce the realizations of a non-ergodic stochastic process, which will enable us to generate the composite texture.

\section*{Acknowledgements}
This study was supported by Hawai'i Experimental Program to Stimulate Competitive Research (EPSCoR) provided by the National Science Foundation Research Infrastructure Improvement (RII) Track-1: ’Ike Wai: Securing Hawai'i's Water Future Award OIA as well as an appointment to the Faculty Research Participation Program at the U.S. Engineer Research and Development Center, Coastal and Hydraulics Laboratory administered by the Oak Ridge Institute for Science and Education through an interagency agreement between the U.S. department of Energy and ERDC.
This work was also supported by the Laboratory Directed Research and Development program at Sandia National Laboratories. Sandia National Laboratories is a multimission laboratory managed and operated by National Technology and Engineering Solutions of Sandia, LLC., a wholly owned subsidiary of Honeywell International, Inc., for the U.S. Department of Energy’s National Nuclear Security Administration under contract DE-NA-0003525. Any subjective views or opinions that might be expressed in the paper do not necessarily represent the views of the U.S. Department of Energy or the United States Government.

\appendix
\section*{Appendix A}
\noindent We added SAGAN architecture and hyperparameters of each test cases used in this paper.
\begin{table}[h]
\centering
\renewcommand\thetable{A.1} 
\caption{Hyperparameters and architecture of the networks used in this paper}
\label{tab:appA}
\resizebox{\textwidth}{!}{%
\begin{tabular}{|c|c|c|c|c|c|c|c|c|c|}
\hline
\multirow{2}{*}{\textbf{Case}} & \multirow{2}{*}{\textbf{Model}} & \multicolumn{3}{c|}{\textbf{TI}} & \multirow{2}{*}{\textbf{Z'}} & \multicolumn{2}{c|}{\textbf{$\sharp$ of filter per layer}} & \multirow{2}{*}{\textbf{\begin{tabular}[c]{@{}c@{}}Generated\\ size\end{tabular}}} & \multirow{2}{*}{\textbf{\begin{tabular}[c]{@{}c@{}}Size of \\ Kernel, Stride\end{tabular}}} \\ \cline{3-5} \cline{7-8}
 &  & \textbf{Name} & \textbf{$\sharp$} & \textbf{Size} &  & \textbf{G model} & \textbf{D model} &  &  \\ \hline
\multirow{8}{*}{\textbf{2D}} & \multirow{2}{*}{DCGAN} & \multirow{4}{*}{Strebelle} & 1 & 256 $\times$ 256 & \multirow{3}{*}{32 $\times$ 32} & \multirow{3}{*}{64-32-16} & \multirow{3}{*}{16-32-64} & \multirow{3}{*}{256 $\times$ 256} & \multirow{8}{*}{\begin{tabular}[c]{@{}c@{}}5 $\times$ 5\\   2 $\times$ 2\end{tabular}} \\ \cline{4-5}
 &  &  & 100 & (128 $\times$ 2) $\times$ (128 $\times$ 2) &  &  &  &  &  \\ \cline{2-2} \cline{4-5}
 & \multirow{2}{*}{SAGAN} &  & 100 & 130 $\times$ 130 &  &  &  &  &  \\ \cline{4-9}
 &  &  & 1 & 256 $\times$ 256 & 33 $\times$ 65 & 128-64-32-16 & 16-32-64-128 & 528 $\times$ 1040 &  \\ \cline{2-9}
 & \multirow{2}{*}{DCGAN} & \multirow{4}{*}{Beadpack} & 256 & 256 $\times$ 256 & 32 $\times$ 32 & \multirow{3}{*}{64-32-16} & \multirow{3}{*}{16-32-64} & 256 $\times$ 256 &  \\ \cline{4-6} \cline{9-9}
 &  &  & 256 & (128 $\times$ 2) $\times$ (128 $\times$ 2) & 64 $\times$ 64 &  &  & 512 $\times$ 512 &  \\ \cline{2-2} \cline{4-6} \cline{9-9}
 & \multirow{2}{*}{SAGAN} &  & 256 & \multirow{2}{*}{256 $\times$ 256} & 32 $\times$ 32 &  &  & 256 $\times$ 256 &  \\ \cline{4-4} \cline{6-9}
 &  &  & 256 &  & 64 $\times$ 64 & 128-64-32 & 32-64-128 & 512 $\times$ 512 &  \\ \hline
\multirow{10}{*}{\textbf{3D}} & \multirow{2}{*}{DCGAN} & \multirow{5}{*}{Beadpack} & 1 & 256 $\times$ 256 $\times$ 256 & 4 $\times$ 4 $\times$ 4 & 64-32-16-8-4-4 & 4-4-8-16-32-64 & 256 $\times$ 256 $\times$ 256 & \multirow{10}{*}{\begin{tabular}[c]{@{}c@{}}3 $\times$ 3 $\times$ 3\\    2 $\times$ 2 $\times$ 2\end{tabular}} \\ \cline{4-9}
 &  &  & 100 & 128 $\times$ 128 $\times$ 128 & 8 $\times$ 8 $\times$ 8 & 64-32-16-8 & 8-16-32-64 & 128 $\times$ 128 $\times$ 128 &  \\ \cline{2-2} \cline{4-9}
 & \multirow{3}{*}{SAGAN} &  & 1 & \multirow{3}{*}{256 $\times$ 256 $\times$ 256} & \multirow{2}{*}{8 $\times$ 8 $\times$ 8} & 64-32-16-8-4 & 4-8-16-32-64 & 256 $\times$ 256 $\times$ 256 &  \\ \cline{4-4} \cline{7-9}
 &  &  & 1 &  &  & 64-32-16-8 & 8-16-32-64 & 128 $\times$ 128 $\times$ 128 &  \\ \cline{4-4} \cline{6-9}
 &  &  & 1 &  & 8 $\times$ 8 $\times$ 10 & 64-32-16-8-4 & 4-8-16-32-64 & 256 $\times$ 256 $\times$ 320 &  \\ \cline{2-9}
 & \multirow{2}{*}{DCGAN} & \multirow{5}{*}{\begin{tabular}[c]{@{}c@{}}Fold \\ Aquifer\end{tabular}} & 1 & 128 $\times$ 128 $\times$ 128 & \multirow{5}{*}{8 $\times$ 8 $\times$ 8} & 64-32-16-8 & 8-16-32-64 & 128 $\times$ 128 $\times$ 128 &  \\ \cline{4-5} \cline{7-9}
 &  &  & 100 & 64 $\times$ 64 $\times$ 64 &  & 64-32-16 & 16-32-64 & 64 $\times$ 64 $\times$ 64 &  \\ \cline{2-2} \cline{4-5} \cline{7-9}
 & \multirow{3}{*}{SAGAN} &  & 1 & \multirow{3}{*}{128 $\times$ 128 $\times$ 128} &  & 64-32-16-8 & 8-16-32-64 & 128 $\times$ 128 $\times$ 128 &  \\ \cline{4-4} \cline{7-9}
 &  &  & 1 &  &  & 64-32-16 & 16-32-64 & 64 $\times$ 64 $\times$ 64 &  \\ \cline{4-4} \cline{7-9}
 &  &  & 1 &  &  & 64-32-16-8-4 & 4-8-16-32-64 & 256 $\times$ 256 $\times$ 256 &  \\ \hline
\end{tabular}%
}
\end{table}


\end{document}